# RoboPainter - A Detailed Robot Design for Interior Wall Painting

Mohamed Sorour, *Student Member, IEEE*

*Abstract*—High demand for painters is required nowadays and foreseen in the near future for both developed and developing countries. To satisfy such demand, this paper presents the detailed computer aided design (CAD) model of a fully functional wall painting robot for interior finishes. The RoboPainter is capable of performing full scale wall-ceil painting in addition to decorative wall drawings. The 8 degrees of freedom (DOF) mobile robot structure consists of a 6DOF spray painting arm mounted on a 2DOF differentially driven mobile base. The design presented endows several achievements in terms of total robot mass and painting rate as compared to existing literature. Detailed dynamic model parameters are presented to allow for further enhancement in terms of robot motion control.

## I. INTRODUCTION

Manual wall-ceil painting is repetitive, time consuming, physically exhaustive, and dangerous. On one hand, all human-health centered drawbacks are in favor of robotizing the application. On the other hand employment of painters is expected to grow by 20% from 2012 to 2022 according to the U.S. bureau of labor statistics [1], which requires an increase of 62,600 in the number of painters during this period in the U.S. alone. At least an equal figure can be expected in the dense populated developing countries, and regions with higher birth rates; for example the North African region, India, China and South America. Despite these facts, no (to the best of the author's knowledge) robotic interior painting device is commercially available on market at the time of writing this article.

Little research work on interior painting finishes can be found in the literature, despite its environmental (saving human effort and health) and commercial significance. Kahane et al. [2] studied human-robot integration in construction site and promising results were obtained. In the case of wall painting, reduction in total painting time of about 70% was reported, which is subject to an increase of 20% if robots are additionally integrated in ceil painting. Warszawsky et al. [3] described a system "TAMIR" for interior finish works. It can perform four tasks; namely wall painting, plastering, tiling, and masonry. The system consists of 6DOF industrial robot arm (GM S-700) mounted on a 3 wheeled mobile robot. In its operation, the system moves to a predefined workstation, deploys its four stabilizing legs and starts its program. However the weight of the arm ($500 kg$) limits, its portability and its indoor domestic use. Naticchia et

M. Sorour is with the Laboratory for Computer Science, Microelectronics and Robotics LIRMM – Université de Montpellier 2 / CNRS, Montpellier, France - PSA Peugeot-Citroen, Velizy Villacoublay, France. sorour@lirmm.fr. This research activity is privately funded by the author, was initially supported by Egypt-Japan University of Science and Technology (E-JUST), New Borg-El-Arab city, Alexandria, Egypt.

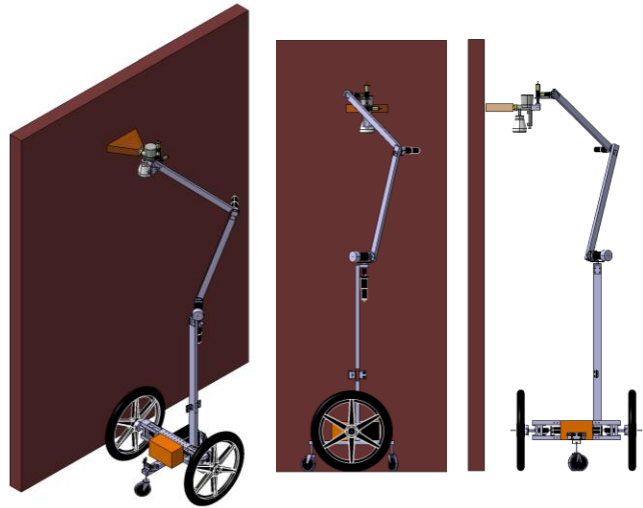

Fig.1 CAD model of the designed RoboPainter

al. [4] developed a scaled down robot setup for wall spray painting together with a multi-color spray gun. The developed robot "Pollock#1" consists of fixed base 6DOF arm with $0.4\ m$ reach, which limits its portability and operability. Aris et al. [5] developed a full scale mechanism for ceil painting. Robot working envelope of $84 \times 72 \times 122\ cm^3$ is reported together with improvement in ceil painting time and cost, where $46\ m^2$ of ceil were painted in $3.5\ hours$. This ceil painting rate was reported as 1.5 times faster than manual painting. Unfortunately the developed system only paints the ceil and no data is given with regards to its mobility. Teoh et al. [6] developed a cable driven robot chassis "PAINTbot" for interior and exterior wall painting. The robot, hanged by cables using two pivots fixed to the wall might be applicable to exterior wall painting, but is not suitable for interior finishes. Sorour et al. [7] developed a 5DOF roller painting mobile robot, consisting of a 2DOF painting arm fitted to a 3DOF mobile platform. The total system weights $35\ kg$ and painting rate of $10\ m^2/hour$ was reported. Abdellatif in [8] further provided robot design considerations as possible extension to the work done in [7].

Among the most tedious and time consuming tasks entitled to the painter comes the wall-ceil painting (applying one or several coatings of paint or stain), performing which is the main goal of this research activity. In this work the author extends the conceptual design recently presented in [9] by introducing the detailed RoboPainter CAD design shown in Fig.1. The spray painting based mobile robot has in total 8DOF, consisting of 6DOF painting arm mounted on a 2DOF differentially driven mobile base. The designed RoboPainter enhances the present literature in terms of lower total robot mass, higher painting rate, smaller occupied volume in addition to the ability to perform decorative drawing as well as wall-ceil painting. Stability is ensured using 4 contact point wheeled mobile robot (WMR), while low cost constraint is taken into account by employing the differential drive system. In order to develop a fully functional, reliable wall painting system, the author provides the dynamic model in the form of detailed robot inertial parameters to the robotics community for further enhancing the proposed motion planning and dynamics control.

This paper is organized as follows; Section II presents the designed RoboPainter description and the foreseen technical data. Section III details the dynamic models of the mobile base and the painting arm. Section IV introduces the proposed robot operation scenario in terms of self-localization, navigation and wall painting, together with operational assumptions. Conclusions are given in section V.

## II. SYSTEM DESCRIPTION

The RoboPainter is an 8DOF mobile painting robot, the designed technical data of which are given in TABLE I. The painting arm carries at its tip a spray painting gun (whose power rating is underlined in the power consumption entry of TABLE I). Several commercially available painting guns (that fit some physical description) can be attached to the arm tip thanks to its general fixation assembly, for instance refer to [10] for sprayers using $220\,V$ mains electricity voltage and [11] for those employing the $110\,V$ mains. In its design, care was taken to choose off the shelf components so as to shorten the development time and to enable easy, fast component replacement in case of failure.

RoboPainter has its joints actuated by maxon DC-geared motors (except for the arm's second joint being actuated by maxon BLDC-geared motor), all of which are equipped by position encoders and torque controlled using ESCON servo controllers. Despite the long reach of the painting arm ($1.29\,m$), its motion is restricted laterally to the width of a maximum of 2 paint strips in either directions from the robot centerline ($=\pm\,0.5\,m$) (refer to Fig. 2.c). In the motion planning phase of the spray-gun path, the wall is divided into two sections namely; the *core* and the *outline* sections. Fig. 2.c shows the maximum allowable painting workspace per fixed mobile base post in the *core section* painting. The objective of such restriction is twofold; 1) It decreases the maximum torque requirement at each joint and consequently we can use lighter-cheaper motors, and 2) It enhances the human-robot safety by decreasing the volume of the space occupied by the robot structure. The painting arm is capable of delivering a strip of paint sizing $0.25\,m$ width by $2.45\,m$ height in an average time of $10\,seconds$. It is designed to have a maximum paint reach of $2.7\,m$ (depicted in Fig. 2.b) which is the usual room height. The RoboPainter shall be able to perform decorative wall drawings by transforming uploaded pictures into paint path while using smaller spray-nozzle. Based on the cost estimate presented in Table I, the return on investment is expected after $1820\,m^2$ equivalent painting jobs (this figure is based on average labor cost of $8.25\,USD/m^2$) [12].

TABLE I. ROBOPAINTER DESIGN TECHNICAL DATA

| $Total\ mass\ (paint\ cup\ filled)$ | $21.5\,kg$ |
|---|---|
| $Painting\ rate$ | $>200\,m^2/hour$ |
| $Power\ Consumption$ | $(485+\underline{110})W$ |
| $Overall\ dimensions\ (folded)$ | $81.5\times54\times112.5\,cm^3$ |
| $Cost\ estimate\ (prototype)$ | 15K USD |

The 2DOF mobile base consists of 2 actuated fixed wheels in addition to 2 passive castor wheels to provide base stability. The fixed wheels are $20"\,BMX$ pneumatic type, which are very common, cost effective and act as vibration damper (vibration levels during normal spray-gun operation can reach $10\,m/s^2$ [10]). The base weights $13\,kg$ thanks to the aluminum based structure and is capable of supporting additional load of $100\,kg$. It is fitted with 10 sonar sensors (ultra-sonic range finders), 6 of which (blue cones in Fig. 2.d) are used explicitly for obstacle (human operator) avoidance, the remaining 4 sonars (yellow and grey cones in the same figure) are used primarily in localization and navigation. Additionally, the mobile base embeds an inertial measurement unit (IMU) to monitor structure vibrations for the purpose of detecting the *empty paint cup* situation.

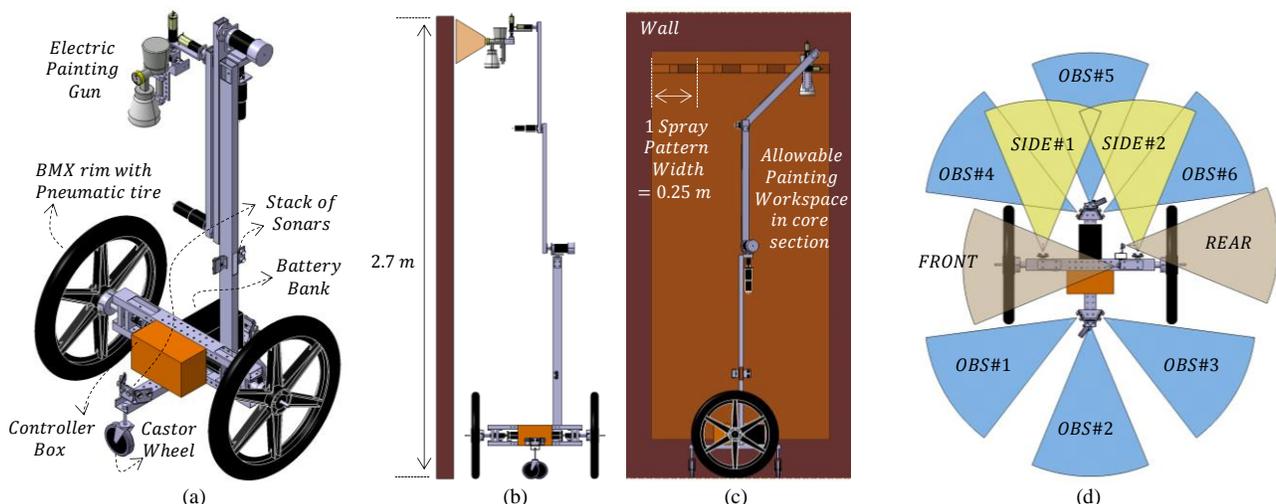

Fig.2 RoboPainter; (a) folded structure with brief description, (b) Fully extended with maximum reach of $2.7\,m$, (c) allowable painting workspace per fixed mobile base post, and (d) mobile base sonar sensors designation with $OBS\#x$ short handing obstacle

## III. DYNAMIC MODEL

In this section we present in brief the dynamic models employed for both the mobile base and the painting arm. Complete kinematic model of the RoboPainter is presented in [9].

### A. Mobile Base Model

Fig. 3.b shows the frames assigned to the links constituting the mobile base; namely the based link, orientable-hub link (orientable link of the castor wheel assembly), the castor wheel link and the fixed wheel link, given the subscripts $b$, $o$, $c$, and $f$ respectively in what follows. The dynamic model of the mobile base subject to non-holonomic constraints is derived using the Euler-Lagrange formalism and is given as (1):

$$M_b(q_b)\ddot{q}_b + C_b(q_b,\dot{q}_b)\dot{q}_b = \Gamma_b + J^T(q_b)\lambda, \quad (1)$$

where $M_b(q_b)$ and $C_b(q_b,\dot{q}_b)$ denote the inertia matrix and the matrix representing the Coriolis and centrifugal effects respectively, $q_b = [\bar{X}_b \quad \beta_c^T \quad \varphi_f^T \quad \varphi_c^T]^T$ is the $9\times 1$ vector of generalized coordinates with $\bar{X}_b = [x_{o_b} \quad y_{o_b} \quad \phi_b]^T$ denoting the position and orientation of the base frame, $\beta_c^T = [\beta_{1c} \quad \beta_{2c}]$ is the vector of castor wheels orientation, $\varphi_f^T = [\varphi_{1f} \quad \varphi_{2f}]$ is the vector of fixed wheel angles and $\varphi_c^T = [\varphi_{1c} \quad \varphi_{2c}]$ is the vector of castor wheel angles. $\Gamma_b$ is the generalized force vector associated with the torques provided by the actuators. $J^T(q_b)\lambda$ represents the generalized forces associated with the kinematic constraints where $J(q_b)$ is the matrix of kinematic constraints and $\lambda$ is the vector of Lagrange multipliers. The configuration kinematic model is given by (2) as follows [9]:

$$\dot{q}_b = S_b(q_b)u_b, \quad (2)$$

where $u_b = [\dot{x}_{o_b}^b \quad \dot{\phi}_b^b]^T$ is the vector of directly controllable degrees of mobility (these velocities are taken with respect to the base frame as indicated by the superscript $b$) and the matrix $S_b$ detailed in [9] spans the null space of $J(q_b)$ satisfying the relation in (3) [13]:

$$J(q_b)S_b(q_b) = 0. \quad (3)$$

From (3) we can eliminate the Lagrange multipliers in (1) by pre-multiplying both sides by $S_b^T(q_b)$ as provided in (4):

$$S_b^T(q_b)M_b(q_b)\ddot{q}_b + S_b^T(q_b)C_b(q_b,\dot{q}_b)\dot{q}_b = S_b^T(q_b)\Gamma_b, \quad (4)$$

Substituting by (2) in (4) we finally obtain the dynamic model of the mobile base in the space of controllable degrees of mobility (5):

$$\widetilde{M}_b(q_b)\dot{u}_b + \widetilde{C}_b(q_b)u_b = S_b^T(q_b)\Gamma_b, \quad (5)$$

with $\widetilde{M}_b(q_b) = S_b^T(q_b)M_b(q_b)S_b(q_b)$ and $\widetilde{C}_b(q_b) = S_b^T(q_b)M_b(q_b)\dot{S}_b(q_b) + S_b^T(q_b)C_b(q_b,\dot{q}_b)S_b(q_b)$.

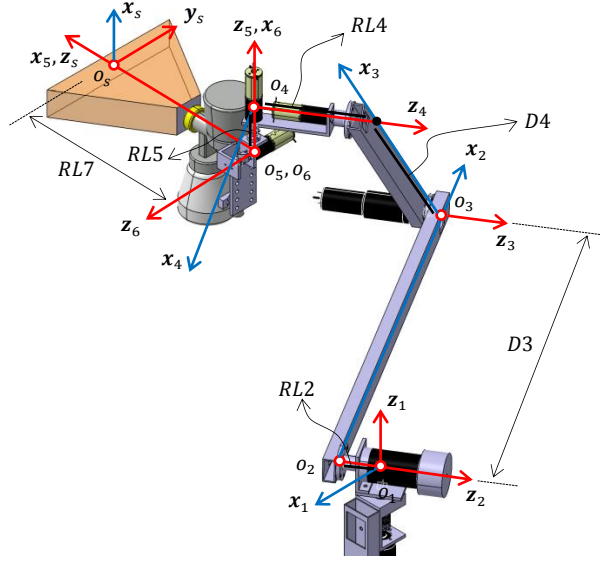

(a) Painting arm model

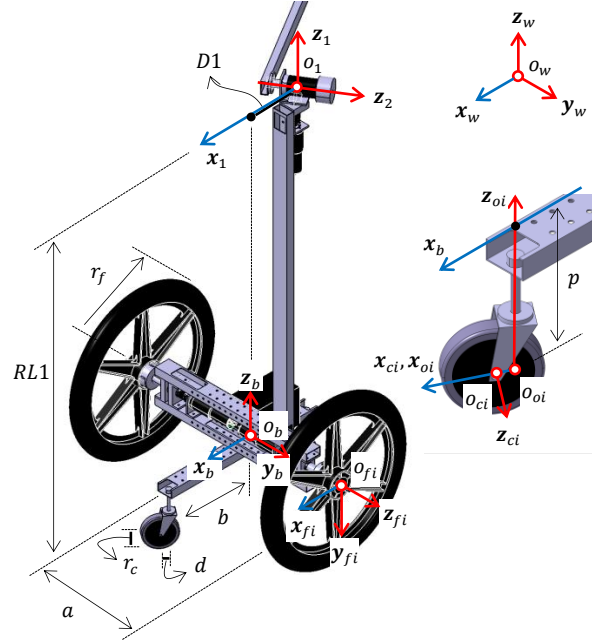

(b) Mobile base model

Fig. 3 RoboPainter geometric parameters and assigned frames for (a) the painting arm and, (b) the mobile base

The inertia matrix $M_b(q_b)$ is obtained from the expression of the total kinetic energy $E_t$ provided in (6) using (7):

$$E_t = E_b + \sum_{i=1}^{2} E_{fi} + \sum_{i=1}^{2} E_{oi} + \sum_{i=1}^{2} E_{ci} + E_a, \quad (6)$$

where $E_b$ and $E_a$ denote the kinetic energy of the base link and the painting arm respectively while $E_{fi}$, $E_{oi}$ and $E_{ci}$ denote the kinetic energy of the $i^{th}$ fixed wheel, orientable-hub and castor wheel respectively.

$$M_{bij} = \frac{\partial E_t}{\partial \dot{q}_{bi} \dot{q}_{bj}}. \qquad (7)$$

In (7) $M_{bij}$ represent the *ij* element of the inertia matrix. The matrix of Coriolis and centrifugal effects $C_b(q_b, \dot{q}_b)$ can then be evaluated using the *Christoffell* symbols $c_{ijk}$ as:

$$C_{bij} = \sum_{k=1}^{9} c_{ijk} \cdot \dot{q}_{bk}, \qquad (8)$$

with,

$$c_{ijk} = \frac{1}{2}\left[\frac{\partial M_{bij}}{\partial q_{bk}} + \frac{\partial M_{bik}}{\partial q_{bj}} - \frac{\partial M_{bjk}}{\partial q_{bi}}\right].$$

### B. Painting Arm Model

Fig. 3.a shows the frames assigned to the painting arm according to the *Khalil-Kleinfinger* notation [14] along with the definition of the arm geometric parameters. In a similar fashion to the methodology presented above, the dynamic model of the painting arm can be formulated as [15]:

$$M_a(q_a)\ddot{q}_a + C_a(q_a,\dot{q}_a)\dot{q}_a + Q(q_a) + \Gamma_{fr} + \Gamma_{ex} = \Gamma_a, \qquad (9)$$

where $q_a$ is the $6 \times 1$ vector of arm generalized coordinates, $Q(q_a)$ represents the required joint torques for gravity compensation, $\Gamma_{fr}$ and $\Gamma_{ex}$ denote vectors of joint torques to overcome friction and external forces respectively. Although spray painting doesn't involve any wall-contact, $\Gamma_{ex}$ exists with significant contribution due to 1) reaction force resulting from spray pushing and 2) vibration of the spray gun during normal operation (acceleration levels can reach $10 \, m/s^2$). The matrices $M_a(q_a)$ and $C_a(q_a,\dot{q}_a)$ can be obtained as described in the mobile base case, while the elements of the vector $Q(q_a)$ are obtained using (10) where $U$ is the expression of the arm's potential energy.

$$Q_i = \frac{\partial U}{\partial q_{ai}}, \qquad (10)$$

A symbolic form of the dynamic model for the painting arm is obtained easily using OpenSYMORO software package [16] by feeding the table of geometric parameters presented in [9]. In order to enable further enhancement in the dynamic modeling and control of the RoboPainter, complete inertial parameters; masses, C.G point coordinates, inertia tensors of each link together with the rotor inertia and motor torque constants for the mobile base and the painting arm are provided in appendix I. The C.G coordinates are taken with respect to the link frames depicted in Fig. 3, while the inertia tensors are computed with respect to a frame parallel to the frames indicated in the same figure.

## IV. OPERATION SCENARIO

Eight sonar sensors fitted to the mobile base are used to localize the robot within the room. The maximum detectable range of each is $5 \, m$ with an accuracy of $1 \, cm$. Such low accuracy is acceptable for the painting operation which requires positioning the paint gun facing the wall with an offset in the range of $10 - 25 \, cm$. Robot operating assumptions can be listed as:

1) Room side-walls are perpendicular to each other,
2) Initial mobile base posture is set such that the front and right sonars are facing different walls,
3) Room geometric model (location of windows, doors, etc.) is known with respect to the world frame to be programmed out of the painting path,
4) Paint cup refill is to be done by human operator,
5) The *spray pattern* is rectangular in shape with an area of $5 \times 26 \, cm^2$,

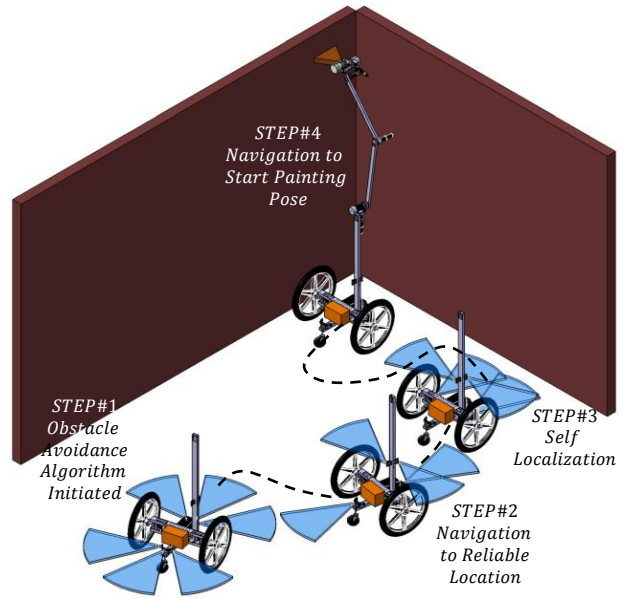

Fig. 4 Robot localization and navigation to the *start painting* pose, the painting arm is omitted for the first 3 steps for visual conciseness

Assumptions 2) and 3) can be relaxed in the future by employing vision system. The designed operation sequence (scenario) can then be summarized as:

1) Obstacle avoidance algorithm using obstacle ($OBS\#1$ - $\#6$) (blue cones in Fig. 2.d) sonar measurements is initiated and maintained till job termination to enhance human-robot safety (see step#1 in Fig. 4, although operational, these 6 sonars will not appear in next steps to avoid visual ambiguity),
2) Move to the *reliable location* (step#2 in Fig. 4), at which the front and right sonars indicate distances within the detectable range ($0.02 - 5m$),
3) Measure the orientation angle using side ($SIDE\#1$ - $\#2$) (yellow cones in Fig. 2.d) sonars (step#3 in Fig. 4),
4) Using above measurements the room's front-right corner is detected and registered as the world frame $\mathcal{R}_w$. The transformation matrix $T_b^w$ is obtained,
5) Self-localization based on dead reckoning will be used afterwards with frequent error corrections using different sonars,

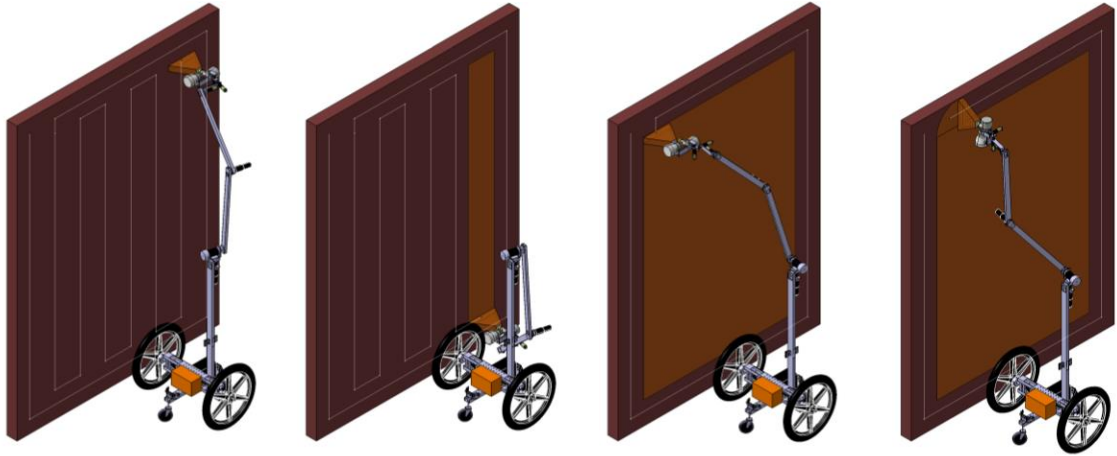

| (a) the whole paint path | (b) vertical paint strip | (c) the core section painted | (d) painting the outline section |

Fig. 5 Wall paint path planning; (a), (b) shows the vertical paint path employed to minimize mobile base motion, (c) and (d) rotating the *roll* angle by 90º to switch from core to outline section painting

6) The robot navigates to the *start painting* pose,
7) Once the robot reaches the *start painting* pose, the painting arm follows a linear with $5^{th}$ order blends motion profile for the point to point path indicated in Fig. 5.a. This high order motion profile is necessary to reduce spray-gun vibration - due to forces induced by infinite acceleration requirement with lower order profiles - especially with such long span structure. Another source of vibration is the spray gun itself during normal operation with acceleration levels in the range $2.5 \sim 10\ m/s^2$ with uncertainty [10].
8) Online paint path planning is done based on the left sonar measurement where the wall is divided into vertical *paint strips* (Fig. 5.a) each of which is $25\ cm$ in width with $1\ cm$ overlap. Vertical Paint path (Fig. 5.b) is employed to minimize mobile base movement,
9) The painting path is divided into 2 sections; namely the core section (refer to Fig. 5.c) and the outline section (Fig. 5.d) due to physical dimensions of available painting guns that oversize the spray pattern. In the outline section, the horizontal paint strip, $25\ cm$ in height left during the core section painting is accomplished where the painting arm changes the *roll* angle of the spray gun by 90º.
10) Having the first wall completely painted, the mobile base rotates counterclockwise and starts the previous sequence again for other remaining walls till the initial robot orientation is detected (first wall painted), at which point the job terminates. Robot operation can be terminated-paused by the user using push buttons.
11) Spray gun vibration is continuously monitored using IMU to detect *empty paint cup* situations, at which the painting program pauses, painting arm pose changes to a prescribed, easy-access configuration for manual *paint refill* by the operator.

## V. CONCLUSION

The detailed design of the RoboPainter, a sprayer based wall painting robot for interior finishes has been introduced. The robot was designed to allow for full scale wall-ceil painting in addition to decorative wall drawings. RoboPainter's dynamic model has been formulated-presented together with the detailed inertial parameters of its constituting links and actuators. Operation scenario of the robot has been discussed along with operational assumptions. Several conceptual objectives for an indoor painting robot such as light weight, high painting rate and fast return of investment has been realized.

## APPENDIX I

This appendix details RoboPainter's inertial parameters and motor technical data require for dynamic simulation and control. In the following tables, the parameter symbols are named with numbers $(1-6)$ indicating the corresponding arm link and with $b$ indicating the base link, $o$ indicating the orientable hub link (part of the castor wheel assembly), $c$ indicating the castor wheel link and $f$ indicating the fixed wheel link. Refer to Fig. 3 for other symbols. For motor data, $Ia1f$ indicate the reflected rotor inertia at the gearbox side for the first fixed wheel and $Kti$ denote the torque constant of the $i^{th}$ geared-motor. For compactness, the units are indicated only for the first table entry.

TABLE II. ROBOPAINTER GEOMETRIC PARAMETERS

| $RL1$ | $1094\ mm$ | $RL4$ | 202.4 |
|---|---|---|---|
| $D1$ | 78.4 | $D4$ | 590 |
| $RL2$ | 64.4 | $RL5$ | 83 |
| $D3$ | 700 | $RL7$ | 307.5 |
| $a$ | 350 | $b$ | 250 |
| $r_c$ | 50 | $r_f$ | 254 |
| $p$ | 204 | $d$ | 22 |

TABLE III. ROBOPAINTER DYNAMIC PARAMETERS – MASS

| $M1$ | 1.509 $kg$ | $M4$ | 0.448 |
|---|---|---|---|
| $M2$ | 0.563 | $M5$ | 0.248 |
| $M3$ | 1.241 | $M6$ | 2.435 |
| $Mb$ | 9.876 | $Mo$ | 0.073 |
| $Mc$ | 0.117 | $Mf$ | 1.984 |

TABLE IV. ROBOPAINTER DYNAMIC PARAMETERS – C.G COORDINATE

| $X1$ | 0 $mm$ | $Y1$ | 33.775 | $Z1$ | 1.946 |
|---|---|---|---|---|---|
| $X2$ | 350 | $Y2$ | 0 | $Z2$ | $-10$ |
| $X3$ | 92.378 | $Y3$ | 0 | $Z3$ | $-76.738$ |
| $X4$ | 0 | $Y4$ | 6.142 | $Z4$ | 53.644 |
| $X5$ | 0 | $Y5$ | 0.965 | $Z5$ | 6.813 |
| $X6$ | $-43.07$ | $Y6$ | 89.26 | $Z6$ | 0.951 |
| $Xb$ | $-15.9$ | $Yb$ | 17.43 | $Zb$ | 107.1 |
| $Xo$ | $-18.78$ | $Yo$ | 0 | $Zo$ | 48.84 |
| $Xc$ | 0 | $Yc$ | 0 | $Zc$ | 0 |
| $Xf$ | 0 | $Yf$ | 0 | $Zf$ | 0 |

TABLE V. ROBOPAINTER DYNAMIC PARAMETERS – INERTIA TENSOR I

| $XX1$ | $7.4 \times 10^{-4}$ $kg.m^2$ | $YY1$ | 0.003 | $ZZ1$ | 0.003 |
|---|---|---|---|---|---|
| $XX2$ | 0.035 | $YY2$ | $1.76 \times 10^{-4}$ | $ZZ2$ | 0.035 |
| $XX3$ | 0.051 | $YY3$ | 0.003 | $ZZ3$ | 0.049 |
| $XX4$ | 0.001 | $YY4$ | 0.001 | $ZZ4$ | $8.86 \times 10^{-5}$ |
| $XX5$ | $1.01 \times 10^{-4}$ | $YY5$ | $8.03 \times 10^{-5}$ | $ZZ5$ | $4.63 \times 10^{-5}$ |
| $XX6$ | 0.004 | $YY6$ | 0.015 | $ZZ6$ | 0.017 |
| $XXb$ | 1.05 | $YYb$ | 0.953 | $ZZb$ | 0.234 |
| $XXo$ | $3.67 \times 10^{-5}$ | $YYo$ | $3.1 \times 10^{-5}$ | $ZZo$ | $2.27 \times 10^{-5}$ |
| $XXc$ | $10^{-4}$ | $YYc$ | $10^{-4}$ | $ZZc$ | $1.92 \times 10^{-4}$ |
| $XXf$ | 0.029 | $YYf$ | 0.029 | $ZZf$ | 0.055 |

TABLE VI. ROBOPAINTER DYNAMIC PARAMETERS – INERTIA TENSOR II

| $XY1$ | $-3.4 \times 10^{-7}$ | $XZ1$ | $-10^{-4}$ | $YZ1$ | $-8.1 \times 10^{-8}$ |
|---|---|---|---|---|---|
| $XY2$ | $1.7 \times 10^{-17}$ | $XZ2$ | $2.6 \times 10^{-18}$ | $YZ2$ | $-7.1 \times 10^{-4}$ |
| $XY3$ | $2.88 \times 10^{-4}$ | $XZ3$ | $-3.7 \times 10^{-5}$ | $YZ3$ | 0.006 |
| $XY4$ | $1.9 \times 10^{-8}$ | $XZ4$ | $-3.7 \times 10^{-8}$ | $YZ4$ | $7.9 \times 10^{-5}$ |
| $XY5$ | $1.5 \times 10^{-8}$ | $XZ5$ | $-2 \times 10^{-8}$ | $YZ5$ | $-1.8 \times 10^{-6}$ |
| $XY6$ | 0.004 | $XZ6$ | $-10^{-4}$ | $YZ6$ | $1.87 \times 10^{-4}$ |
| $XYb$ | 0.009 | $XZb$ | 0.064 | $YZb$ | $-0.124$ |
| $XYo$ | 0 | $XZo$ | $7.742 \times 10^{-6}$ | $YZo$ | 0 |
| $XYc$ | 0 | $XZc$ | 0 | $YZc$ | 0 |
| $XYf$ | 0 | $XZf$ | 0 | $YZf$ | 0 |

TABLE VII. ROBOPAINTER MOTORIZATION DATA

| $Ia1$ | 0.0391 $kg.m^2$ | $Kt1$ | 3.858 $N.m/A$ |
|---|---|---|---|
| $Ia2$ | 6.402 | $Kt2$ | 12.283 |
| $Ia3$ | 0.0391 | $Kt3$ | 3.858 |
| $Ia4$ | 0.0414 | $Kt4$ | 8.464 |
| $Ia5$ | 0.0414 | $Kt5$ | 8.464 |
| $Ia6$ | 0.0414 | $Kt6$ | 8.464 |
| $Ia1f$ | 0.0391 | $Kt1f$ | 3.858 |
| $Ia2f$ | 0.0391 | $Kt2f$ | 3.858 |